\documentclass{styles/svproc}
\usepackage{url}
\usepackage{amsfonts}
\usepackage{amsmath}
\usepackage{float}
\usepackage{amssymb}
\usepackage{xcolor}
\usepackage{algorithmic}
\usepackage{graphicx}
\usepackage{booktabs}
\usepackage{multirow}
\usepackage{array}
\usepackage{soul}
\usepackage[a4paper, left=133pt, right=113pt, top=91pt, bottom=176pt]{geometry}
\soulregister\cite7

\newfloat{algorithm}{t}{lop}

\definecolor{highlander_blue}{RGB}{0,61,165}

\makeatletter
\newcommand\fs@spaceruled{\def\@fs@cfont{\bfseries}\let\@fs@capt\floatc@ruled
  \def\@fs@pre{\vspace{0.5\baselineskip}\hrule height.8pt depth0pt \kern2pt}%
  \def\@fs@post{\kern2pt\hrule\vspace{-0.95\baselineskip}}
  \def\@fs@mid{\kern2pt\hrule\kern2pt}%
  \let\@fs@iftopcapt\iftrue}
\makeatother

\begin{document}
\mainmatter              
\title{Mobile Multi-Robot Navigation under Runtime Uncertainty via Koopman Operator Learning and Nonlinear Model Predictive Control}
\titlerunning{Multi-Robot Koopman-based Navigation}  
%
\author{Xiaobin Zhang \and Konstantinos Karydis
}
%
%
\tocauthor{Xiaobin Zhang and Konstantinos Karydis}
\institute{University of California, Riverside, Riverside CA 92551, USA\\
\email{xzhan548@ucr.edu}, \email{karydis@ucr.edu}
}

\maketitle              

\begin{abstract}
In this work, we developed a nonlinear model predictive control (NMPC) framework that employs learned dynamics via the Koopman Operator theory for mobile multi-robot navigation. 
We formulated and solved NMPC problems using a lifted bilinear Koopman-based model that accurately predicts affine input systems affected by perturbations and uncertainties. 
Two exemplary multi-robot navigation problems are considered: target reaching and formation control. 
The output of our method enables closed-loop multi-robot navigation and formation control in environments populated with obstacles, whereby the Koopman operator-based model used in the NMPC formulation addresses runtime uncertainties, namely, various degrees of random wheel slipping. 
We validated the effectiveness of our method for both problems via extensive numerical simulations in different environments with wheeled robots affected by different amounts of slip and without knowledge of their true dynamic models.
\end{abstract}
\section{Introduction}
The Koopman operator has emerged as a research tool for robot learning~\cite{shi2026koopman}. 
It learns nonlinear dynamics by lifting the variables in the state space into observables in the Koopman space, where the dynamics is considered linear but infinite-dimensional~\cite{williams2015data}. 
Compared to other data-driven control methods, it is far less data-intensive and efficient to train and deploy. 
Deep learning methods leverage vast amounts of data to create models that can
predict and adjust to various operational conditions~\cite{karoly2020deep}. (Deep) reinforcement learning enables robots to learn optimal control policies through trial and error~\cite{oikawa2021reinforcement}, whereas large language models can help interpret complex instructions and adapt to changes in the environment~\cite{chen2024llm}. 
Despite their demonstrated
strengths, they still lack the capacity to be deployed on edge devices mounted on robots, while requiring vast datasets for training~\cite{zhao2024deep}. 
Hence, Koopman-based methods are particularly appropriate for hardware with lower computational capabilities and requiring a comparatively high control frequency~\cite{shi2026koopman}. 
One such application is the real-time navigation of mobile robots in settings affected by uncertain or unknown dynamics.

Several different efforts have employed the Koopman operator in support of mobile robot navigation.  
Notable examples include the modeling and control of a tail-actuated robotic fish~\cite{mamakoukas2021derivative}, micro-aerial vehicles trajectory control~\cite{shi2020data}, spherical robots dynamics estimation~\cite{abraham2017RKexample}, and soft systems model extraction~\cite{bruder2019nonlinear}. 
Crucially, Koopman-based models have been introduced within Model Predictive Control (MPC) schemes (e.g., EDMDc~\cite{korda2018linear} and KRONIC~\cite{kaiser2021data}) as well as Nonlinear MPC (NMPC)~\cite{folkestad2021koopman}. 
Further, several methods~\cite{yu2022autonomous,wang2024k} reduce the complexity of the Koopman-NMPC formulation by linearizing them in order to achieve a higher control frequency. 
Deep learning-based methods~\cite{abtahi2026deep,yi2024adaptive} handle uncertainties by learning the Koopman operator and the associated invariant subspace from data. 
Deep Koopman has also been integrated with Rapidly-exploring Random Trees (RRT) for collision-aware motion planning of space manipulators~\cite{chen2025dk}.

This work focuses on multi-robot learning-based control for navigation. 
On this topic, the performance of traditional methods faces challenges. 
Model-based strategies struggle in highly dynamic scenarios because traditional methods require precise knowledge of system dynamics and often rely on local linearizations~\cite{jelodar2025koopman}. 
Conversely, although pure data-driven machine learning-based methods capture highly nonlinear system dynamics well, the lack of their interpretability, which serves as the basis for system stability analysis, creates a gap for a control framework that takes advantage of both data-driven precision and mathematical bounds~\cite{bruton2019data}. 
We use the Koopman-NMPC framework to fill the gap.
Several previous related works have considered the leader-follower design in two-robot systems, where the dynamics of the leader is known a-priori and the follower dynamics is learned through the Koopman operator. Specifically, Stackelberg game theory has been used with the Koopman operator to guide the trajectory for multi-robot systems~\cite{zhao2024stackelberg}; the operator has also been utilized for formation control of nonholonomic mobile robots under denial-of-service attacks~\cite{zhan2023koopman}, trajectory estimation in unknown nonlinear manifold~\cite{wang2023trajectory}, and data-driven distributed learning of multi-agent systems~\cite{nandanoori2021data}. 
These works only consider learning follower dynamics. 
Distributed consensus-based control law is integrated with graph theory~\cite{jelodar2025koopman} where all agents are considered to have the same dynamics and the Koopman operator learns one robot's nonlinear and nonholonomic behaviors. 
Research on more general multi-robot systems that extend beyond a leader-follower design has used manually engineered utility functions to describe system objectives~\cite{tao2023koopman,zhao2023koopman}. 
However, these works did not consider collision avoidance or address the presence of uncertainty in the system. 

To address these challenges, in this work, we develop a Koopman-NMPC framework for closed-loop multi-robot navigation in obstacle-cluttered environments under runtime uncertainty in the form of stochastic disturbances (wheel slip). 
In contrast to previous Koopman-based multi-robot works, we formulate and solve an optimization problem by considering multiple robots as a whole and learning all robot dynamics simultaneously. 
We also validate our method with two simulation tasks, target reaching and formation control, both with different levels of runtime uncertainty. 
Specifically, 
\begin{itemize}
    \item We extend EDMDc~\cite{korda2018linear} in a bilinear system model framework to learn a multi-robot system's dynamics as a whole. This reduces the complexity and saves both training and deployment time.
    \item Different from existing Koopman-based MPC/NMPC
    methods that consider system dynamics without uncertainty, we consider different levels of slip as uncertainty
    , and demonstrate the system's performance in different environments populated with isolated obstacles.
    \item The whole system is first tested in simulation with differential drive robot models with slip on target reaching and is then assessed in formation control simulations with different levels of uncertainty as well. The proposed method outperforms both the nominal NMPC and the consensus-based method.
\end{itemize}

\section{Technical Background}

Consider the uncertain control-affine dynamical system 
\begin{equation}\label{eq:control-affine}
{\mathop x\limits^. (t) = h_0(x(t)) + \sum_{i=1}^mh_i(x(t ))u_i(t)}\;,
\end{equation}
where $ x\in \mathbb{R}^{n}$ is the state vector 
and $ u\in \mathbb{R}^{m}$ is the forcing input. 
Function 
$h_0$ corresponds to the drift dynamics, whereas $h_i$ are partially known but assumed continuous-differentiable functions.\;\footnote{In this formulation, the control and (unknown) perturbation vector fields are merged. It can be derived directly from the ``canonical'' control-affine dynamical system, where the control and perturbations are added separately to the drift part. The formulation here is more amenable to integration into a Koopman-based framework.} 
This control-affine system is general and can capture mobile robot dynamics~\cite{zhou2023safe}.

\subsection{Koopman Operator Synopsis}
For an unforced dynamical system $\dot{x}=f(x)$ with flow map $\Phi(t,x)$ and observables $\varphi \in \mathcal{F}$, the continuous time Koopman operator $\mathcal{K}:\mathcal{F}\to  \mathcal{F}$ is defined as 
$(\mathcal{K}\varphi)(\cdot)=\varphi \circ \Phi(t,\cdot)$. 
The Koopman operator is characterized by its eigenvalues and corresponding eigenfunctions. 
A function $\phi$ is an eigenfunction of $\mathcal{K}$ if $(\mathcal{K}\phi)(.)=e^{\lambda t}\phi(.)$, where $\lambda$ is the corresponding eigenvalue~\cite{nathan2018applied}. 

The infinitesimal generator of the Koopman operator is given by 
$\displaystyle \lim_{t \to \infty }\frac{\mathcal{K}-I}{t}=\nabla f = \mathcal{L}_f$, 
where $\mathcal{L}_f$ is the Lie derivative with respect to $f$, and satisfies the eigenvalue equation, that is $\mathcal{L}_f\phi=\lambda\phi$. 

Hence, the time-varying observable $ \psi(t, x) = \mathcal{K}\phi(x)$ is the solution~\cite{surana2016koopman} of the PDE
\begin{equation}\label{eq:pde}
\begin{array}{c}
\frac{\partial \psi}{\partial t}= \mathcal{L}_f \psi \\
\psi(0, x) = \phi(x_0)
\end{array}\;,
\end{equation}
where $x_0$ is the initial condition of the unforced system. 

A vector-valued observable $\mathrm{g}(\cdot)$ may be expressed in terms of Koopman eigenfunctions $\phi_i$ as $\mathrm{g}(\cdot) = \sum_{i=1}^{\infty }\phi_i(\cdot)\mathrm{v}_i$, where $\mathrm{v}_i$ denotes the Koopman modes of observable $\mathrm{g}(\cdot)$. 
The Koopman modes are obtained from the projection of the observable on the span of the Koopman eigenfunctions. 
Exploiting the linearity of $\mathcal{K}$ and substituting the eigenfunction definition $(\mathcal{K}\phi_i)(\cdot) = e^{\lambda_i t}\phi_i(\cdot)$, we obtain:
\begin{equation*}\label{eq:koopman_def}
\mathcal{K}\mathrm{g}(\cdot) =\mathcal{K} \sum_{i=1}^{\infty }\phi_i(\cdot)\mathrm{v}_i 
=  \sum_{i=1}^{\infty } \mathcal{K}\phi_i(\cdot)\mathrm{v}_i
= \sum_{i=1}^{\infty } \lambda_i\phi_i(\cdot)\mathrm{v}_i\;.
\end{equation*}
The Koopman eigenvalues and eigenfunctions are properties of the dynamics, whereas the Koopman modes depend on the observables. 

\subsection{Expressing into a Bilinear Koopman Model}
Per \eqref{eq:pde}, $\psi(t, x)$ provides the temporal evolution of an observable $\varphi(\cdot)$ along the trajectory. Applying~\eqref{eq:pde} to the forced system~\eqref{eq:control-affine} leads~\cite{goswami2017global} to the PDE 
\begin{equation}\label{eq:bil_inf}
\frac{\partial \psi}{\partial t}=\mathcal{L}_{h_0}\psi+\sum_{i=1}^{m}u_i\mathcal{L}_{h_i}\psi\;,
\end{equation}
where $\mathcal{L}_{h_i}=h_i \cdot \nabla$ are the corresponding Lie derivatives, and hence are linear operators on the space of $\psi(\cdot)$.
The infinite-dimensional system~\eqref{eq:bil_inf} can be reduced to the finite-dimensional bilinear model 
\begin{equation}\label{eq:bil_fin}
\dot{z}(t) = Az(t) +\sum\limits_{i=1}^{m}N_{i}z(t)u_{i}(t)\;.
\end{equation}
To project the $\mathcal{L}_{h_i}$ operators on the finite-dimensional space, we need to choose a suitable functional basis $\varphi=\{\varphi_1, \varphi_2, \cdots, \varphi_M\}$ in terms of observables. 
We can identify matrices
$A$, $N_i$ as well as the dictionary functions $\varphi$ as 
\begin{equation}\label{eq5}
\left\{\begin{array}{l}
z=\varphi(x)\\
\dot{z} = Az +\sum\limits_{i=1}^{m}N_{i}zu_{i}\\
x=Cz
\end{array}\right.\;.
\end{equation}
Model~\eqref{eq5} can be used to predict the dynamics of $x(t)$ along new trajectories, given its history. 
Here, state $x$ in the Cartesian space is first projected into observable $z$ in the Koopman space, where matrices $A$, $N$ and the control vector $u$ govern the evolution of this observable. 
The observable 
$z$ is then projected back into Cartesian space via matrix $C$.

The bilinear system~\eqref{eq5} can be exactly transformed using the Kronecker product notation into 
$\dot{z}(t)= Az(t)+N(u(t)\otimes z(t))$~\cite{bichiou2018time}. 
Adding a linear input term $B$ to~\eqref{eq5} can enhance the accuracy of the EDMDc algorithm~\cite{kaiser2021data} we consider herein to estimate the Koopman operator from data. 
Thus, the bilinear model to be used in the rest of the paper is
\begin{equation}\label{eq6b}
\dot{z}(t)= Az(t)+N(u(t)\otimes z(t))+Bu(t)\;.  
\end{equation}
Given a collection of sampled time histories of inputs $U=[u_1,\cdots,u_{M-1}]$ and observations $Z=[z_1,\cdots,z_{M-1}]$, matrices $A$, $N$, and $B$ can be obtained as the best bilinear predictor in the lifted space in a least-squares sense, that is 
\begin{equation}\label{eq7}
\underset{A,N,B}{min}\left\| Z'-AZ-N(U\otimes Z)-BU\right\|\;,
\end{equation}
where $Z'=[z_2,\cdots,z_{M}]$. 
Matrix $C$ is computed similarly, 
\begin{equation}\label{eq8}
\underset{C}{min}\left\| X-CZ\right\|\;,
\end{equation}
where $X=[x_1,\cdots,x_{M-1}]$ are the measurements in the state space. 
The analytical solutions of~\eqref{eq7}  and~\eqref{eq8} are
\begin{equation}\label{eq9}
\left\{\begin{array}{l}
[A,N,B]=Z'[Z,U\otimes Z,U]^+ ~~\textrm{and}\\
C=XZ^+
\end{array}\right.\;,
\end{equation}
respectively; $(\cdot)^+$ denotes the Moore–Penrose pseudoinverse.

\section{Problem Formulation and Solution}
We focus on two types of multi-robot navigation problems: (1) target reaching and (2) formation control. 
Different robots in a team are affected by different degrees of uncertainty, taken here to be wheel slipping (defined properly in Section~\ref{sec:results}), while navigating to meet the objectives of the two aforementioned problems. 
The ``nominal" case is the one in which there is no wheel slipping. 

\subsection{Multi-robot Target Reaching} \label{sec3.1}
The goal in this setup is for multiple wheeled robots to reach the desired poses while avoiding collisions with obstacles or other robots and following a smooth trajectory. 
We do not assume knowledge of the nonlinear dynamics model of each robot and utilize the Koopman operator to estimate a model
and, in turn, use it within an NMPC structure to solve this problem. 
In our approach, we stacked the states of each robot as a large state vector. 
The advantage of this approach over learning and controlling individual robots is twofold. 
The larger Koopman matrices help quickly capture common features among the robots and share useful information between them so that they can adapt to the environment more efficiently. 
It also facilitates collision avoidance between robots in the MPC structure because their 
trajectories are planned simultaneously at each time step. 

To solve this Koopman-NMPC problem we formulate it  
\begin{equation}\label{eq10}
\begin{array}{l}
\hspace{1.25cm} \underset{U_t,X_t}{min}\quad  J_T (x,u)=\sum_{t=0}^{T-1} \ell(x_{t+1},u_t)\\
\textrm{subject to:} \\
\begin{array}{l}
f_{eq}(X_t,U_t)=0\\
x(0)=x_0\\
u_t \in \mathbb{U}, \quad  \forall t \in [0,T-1]\\
x_t \in \mathbb{X}, \quad  \forall t \in [0,T]\\
-\left\| p_{it}-p_{oi}\right\|+r_i+r_{oi}\leqslant 0 \quad \forall i \in [1,N_r]\\
-\left\| p_{it}-p_{jt}\right\|+r_i+r_j\leqslant 0, \quad \forall i,j \in [1,N_r], i \neq j
\end{array}
\end{array}
\end{equation}
%
%
where $p_{oi}$ and $r_{oi}$ represent the coordinates and radius of the $i$-th obstacle, respectively, while $p_{it}$ and $p_{jt}$ are the coordinates of the $i$-th and $j$-th robot at time $t$, respectively. 
The robots are represented as circles; $r_i$, $r_j$ are the radii of the $i$-th and $j$-the robots. \;\footnote{This can be applied in practice no matter the footprint of the robot (i.e. circular or rectangular/polygonal) by computing the circle circumscribing its footprint.} 
$N_r$ denotes the total number of robots. 
In problem~\eqref{eq10}, $f_{eq}$ contains dynamic constraints along the prediction horizon $T$. 
These are given by
\looseness=-1
\begin{equation}\label{eq11}
f_{eq}=\begin{bmatrix} x_{t|t}-x_t\\
x_{t+1|t}-f_b(x_{t|t},u_{t}) \\
 \vdots \\
x_{t+T|t}-f_b(x_{t+T-1|t},u_{t+T-1}) \\
\end{bmatrix}\;,
\end{equation}
in which $x_{t+i|t}$ denotes the state at time instant $t+i$ predicted at time instant $t$. $x_t$ and $u_{t}$ denote the current state and control input at time $t$, respectively. 
The Koopman-bilinear prediction handling system uncertainty is 
$f_b(x_{t+i|t},u_{t+i}) = C(A\varphi(x_{t+i|t})+N( u_{t+i}\otimes\varphi(x_{t+i|t}))+Bu_{t+i})$, 
where matrices $A,N,B$ and $C$ are computed via~\eqref{eq9}. 
The objective function to be minimized is chosen as
\begin{equation}\label{e12}
\ell(x_{t+1},u_t) = \left\|d(x_{t+1},x_r)\right\|_{Q}^{2} + \left\| u_t\right\|_{R}^{2}\;,
\end{equation}
where $d(x_{t+1},x_r)$ is the distance between the robots' current state at time $t+1$ and their desired targets, defined as 
$d(x_{t+1},x_r)= [p_{t+1}-p_r,\cos{\theta_{t+1}}-\cos{\theta_{r}}, \sin{\theta_{t+1}}-\sin{\theta_{r}}]^T$, 
where $p_{t+1}$ and $p_r$ are the coordinates of the robots at time $t+1$ and their target positions, and $\theta_{t+1}$ and $\theta_{r}$ are the yaw angles of the robots at time $t+1$ and their target values, respectively\;\footnote{We use a composite distance metric in $SE(2)$ that fuses position in $\mathbb{R}^2$ with the complex number representation of $SO(2)$ for the yaw angle~\cite{lavalle2006planning}. Such a metric has been found appropriate for target reaching and path tracking in the navigation of nonholonomic mobile robots~\cite{karydis2016navigation}.}
Matrices $Q \in \mathbb{R}^{4N_r \times 4N_r}$ and $R \in \mathbb{R}^{2N_r \times 2N_r}$ are real symmetric positive-semidefinite, designed for the nominal problem (i.e. no uncertainty). This formulation can also be extended to time-varying reference signals $\{x_{rk},u_{rk}\}$ for trajectory tracking control. 

Algorithm~\ref{alg1} summarizes the proposed Koopman NMPC 
method. The Koopman operator is first trained offline by solving~\eqref{eq9} with data collected in advance, and then the learned bilinear model is leveraged online to update the NMPC at each time step until the robots converge or any collision with an obstacle or other robots occurs. 
Similar joint offline-online schemes have been found appropriate for Koopman-based control and estimation of prediction error~\cite{shi2021enhancement,shi2022online}.
\floatstyle{spaceruled}
\restylefloat{algorithm}
\begin{algorithm}[!t]
\caption{Koopman-NMPC for Target Reaching}\label{alg1}
 \begin{algorithmic}[1]
 \renewcommand{\algorithmicrequire}{\textbf{Offline:}}
 \renewcommand{\algorithmicensure}{\textbf{Online:}}
 \REQUIRE Compute $A,B,C,N$ via solving~\eqref{eq9}
 \ENSURE Set nominal $Q,R,T$ 
  \FOR {$t = 1,2,\cdots,T$ }
  \STATE $x_0 = x_t$
  \STATE solve (\ref{eq10})
  \STATE $u_t=U_0^*$
  \STATE $x_{t+1}=f_b(x_t,u_t)$
   \ENDFOR
 \end{algorithmic} 
\end{algorithm}

\subsection{Multi-robot Formation Control}
In the second problem we setup a more complex optimization. 
The robots must remain in a desired formation within a leader-follower design. 
They must reach their desired poses, and in addition, follow the designated leader's movement in every step to maintain their formation. 
As in the first problem, the robots must follow smooth trajectories while avoiding obstacle and self-collisions. 
Note that because it is sometimes impossible to achieve the objective with the obstacles, we allow a small amount of deviation from the expected follower trajectories to relax the optimization locally and enable them to circumvent the obstacles. 
No prior knowledge of the nonlinear dynamics of each robot is available, and we leverage the Koopman operator within the NMPC to solve this problem.

The optimization problem to solve in this case is 
\begin{equation}\label{eq13}
\begin{array}{l}
\hspace{1.25cm} \underset{U_t,X_t}{min}\quad  J_T (x,u)=\sum_{t=0}^{T-1} \ell(x_{t+1},u_t)\\
\hspace{0.3cm}\textrm{subject to:} \\
\hspace{0.25cm}
\begin{array}{l}
f_{eq}(X_t,U_t)=0\\
x(0)=x_0\\\
\begin{array}{ll}
\hspace{-0.15cm}
lb \leq x_{fi,t+1}-x_{fi,t} - x_{l,t+1}+x_{l,t} \leq ub,&  \forall t \in [0,T-1], \forall i \in [1,N_r-1]
\end{array}\\
u_t \in \mathbb{U}, \quad  \forall t \in [0,T-1]\\
x_t \in \mathbb{X}, \quad  \forall t \in [0,T]\\
-\left\| p_{it}-p_{oi}\right\|+r_i+r_{oi}\leqslant 0 \quad \forall i \in [1,N_r]\\
-\left\| p_{it}-p_{jt}\right\|+r_i+r_j\leqslant 0, \quad \forall i,j \in [1,N_r], i \neq j
\end{array}
\end{array}
\vspace{-5pt}
\end{equation}
where $x_{fi,t}$ and $x_{fi,t+1}$ are the $i$th follower robot's states, and $x_{l,t+1}$ and $x_{l,t}$ are the leader robot's states, at times $t$ and $t+1$, respectively. $lb$ and $ub$ are the lower and upper bounds of the trajectory deviation.
The other variables and the formulation of the objective function $\ell(x_{t+1},u_t)$ are defined as in Section~\ref{sec3.1}. 
Matrix $Q$ is revised to reflect the different importance of the leader and followers. 
The distances between each robot and its corresponding destination were set to be the same to ensure that maintaining the formation was feasible.
Algorithm~\ref{alg2} summarizes the method. 
The data used to train the system are as in Section~\ref{sec3.1}. 

\floatstyle{spaceruled}
\restylefloat{algorithm}
\begin{algorithm}[!t]
\caption{Koopman-NMPC for Formation Control}\label{alg2}
 \begin{algorithmic}[1]
 \renewcommand{\algorithmicrequire}{\textbf{Offline:}}
 \renewcommand{\algorithmicensure}{\textbf{Online:}}
 \REQUIRE Compute $A,B,C,N$ via solving~\eqref{eq9}
 \ENSURE Set nominal $Q,R,T$ 
  \FOR {$t = 1,2,\cdots,T$ }
  \STATE $x_0 = x_t$
  \STATE solve (\ref{eq13})
  \STATE $u_t=U_0^*$
  \STATE $x_{t+1}=f_b(x_t,u_t)$
   \ENDFOR
 \end{algorithmic}
\end{algorithm}

\section{Results}\label{sec:results}

We first assessed the state prediction for a 3-robot system with differential drive with slip dynamics models. 
Then, with the Koopman operator trained using the collected data, we tested the efficacy of the Koopman-NMPC in nominal conditions and under the effect of uncertain slipping velocities. 
We also compared the proposed method with the nominal NMPC and consensus-based methods~\cite{liu2020consensus}. 
All experiments in this paper were performed on a desktop computer with an AMD R5-7600X CPU at 5.2GHz and 32GB of DRAM at 6000MHz.

\subsection{Differential-drive Mobile Robot System Model with Slip}\label{4.1}

In the first part, we employed a differential-drive model with slip for each robot. 
The governing equations~\cite{ryu2011differential} are
\begin{equation}\label{eq16}
\left\{\begin{array}{l}
\dot{x}_i=(v_i+v_{ti})\cos\theta_i +v_{si} \sin\theta_i \\
\dot{y}_i=(v_i+v_{ti})\sin\theta_i -v_{si} \cos\theta_i \\
\dot{\theta}_i=\omega_i +\omega_{si}  \\
\end{array}\right.\;, \quad i = 1,2,3,
\end{equation}
where $v_i$ and $\omega_i$ are the linear and angular velocities of the $i$-th robot, $v_{ti}$ and $v_{si}$ represent the slip velocities in the forward direction and normal to the forward direction for the $i$-th robot, respectively, and $\omega_{si}$ is the angular slip velocity of the $i$-th robot. We assume that slip velocities are given by
\begin{equation}\label{eq17}
v_{si} = \alpha_i  v_i, 
\quad v_{ti} = \beta_i  v_i,
\quad \omega_{si} = \gamma_i  \omega_i\;.
\end{equation}
Random variables $\alpha_i$, $\beta_i$, $\gamma_i$ are exponentially distributed,
\begin{equation}\label{eq15}
\alpha_i \sim \text{Exp}(\lambda_i),\; \beta_i \sim \text{Exp}(\lambda_i),\; \gamma_i \sim \text{Exp}(\lambda_i)\;,
\end{equation}
where $\lambda_i$ is the mean of each distribution (termed as the slipping coefficient in the remainder of this paper). 
We consider the Koopman-bilinear model obtained using the following settings. 
The minimalist\;\footnote{The basis of the functions used here is a subset of the more general dictionary based on Hermite polynomials and trigonometric functions in~\cite{shi2021acd}.} dictionary $\varphi(x)=[1,p_{1}, \cos \theta_1, \sin\theta_1, p_{2},\\ \cos \theta_2, \sin\theta_2, p_{3}, \cos \theta_3, \sin\theta_3]$ was deployed. 
The scaled dynamics were discretized using the fourth-order Runge–Kutta (RK4) method with period $T_s = 0.1 s$. 
We simulated $1000$ trajectories over $1000$ sampling periods. 
The control input for each robot in each trajectory was a random signal uniformly distributed on the unit box $[-1, 1]^2$. The initial states of the robots for all trajectories were generated randomly with a uniform distribution over $[-10, 10]^3$.
The Koopman prediction accuracy is quantified by the root mean squared error, $RMSE(\%)=100\sqrt\frac{{\sum_{t=1}^{T}(x_t-\hat{x}_t)^2}}{T}$, where $\hat{x}_t$ is the prediction of the robot state at time $t$.

\begin{figure}[!t]
\vspace{2pt}
\centering
\includegraphics[trim={0.05cm, 0.5cm, 1.15cm, 1.25cm},clip,width=0.35\textwidth]{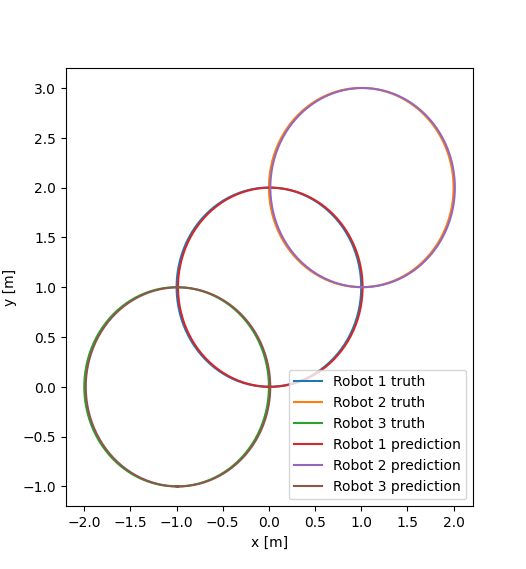}
\includegraphics[trim={0.05cm, 0.5cm, 1.15cm, 1.25cm},clip,width=0.35\textwidth]{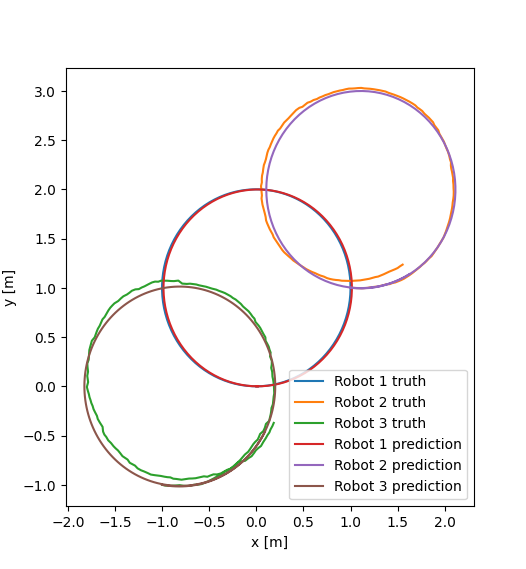}
\vspace{-13pt}
\caption{Comparison of the Koopman-bilinear approximation with the exact differential-drive robot model without (left) and with (right) slip.}\label{fig:model}
\vspace{-5pt}
\end{figure}

To validate the Koopman-based state estimation, we first tested the nominal robot model by setting the following parameters: $v=0.3m/s$, \quad $\omega=0.3rad/s$, $\alpha=\beta=\gamma=0$, for each robot, which should make the robots go in a circle, and the Koopman operator predicts a trajectory with a length of 210 steps in the future. 
The starting states of the robots are $[0,0,0]$, $[-1,-1,0]$ and $[1,1,0]$. 
Figure~\ref{fig:model} (left) confirms that the Koopman-bilinear model can approximate the exact differential-drive robot model. 
The RMSE is evaluated to be $[2.12\%,2.10\%,2.08\%]$ for robots $1,2,3$, respectively, with the above fixed parameters.\;\footnote{This accuracy is dependent on the training data dimension and the sampling time. For instance, the RMSE increases to [$3.96\%,3.96\%,4.29\%$] for $500$ trajectories, $500$ sampling periods, and $T_s=0.2s$.}

\begin{figure*}[!t]
\vspace{6pt}
\centering
\includegraphics[trim={0.05cm, 0.49cm, 0.5cm, 1.65cm},clip,width=0.325\textwidth]{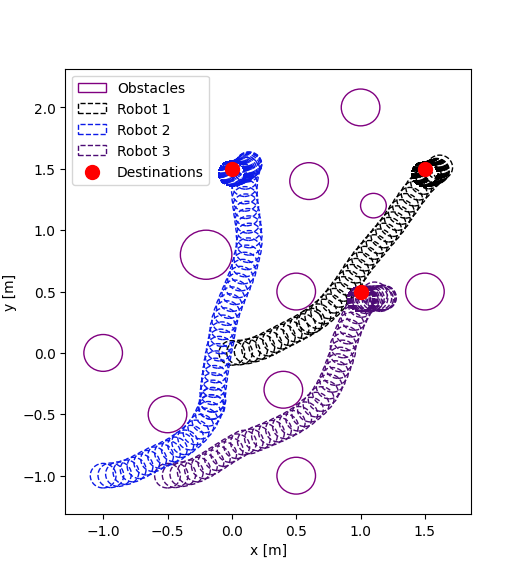}
\includegraphics[trim={0.05cm, 0.49cm, 0.5cm, 1.65cm},clip,width=0.325\textwidth]{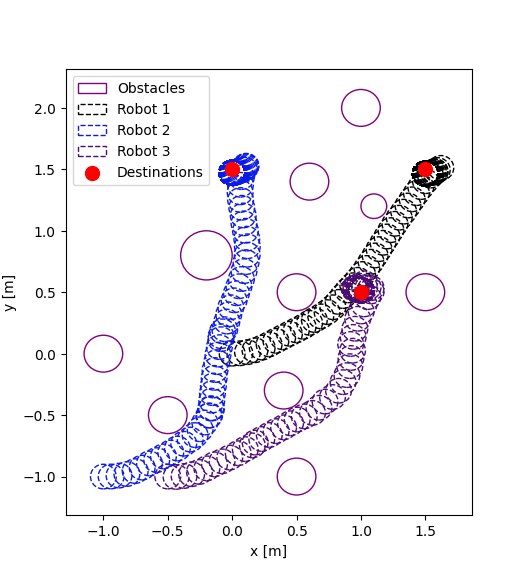}
\includegraphics[trim={0.05cm, 0.49cm, 0.5cm, 1.65cm},clip,width=0.325\textwidth]{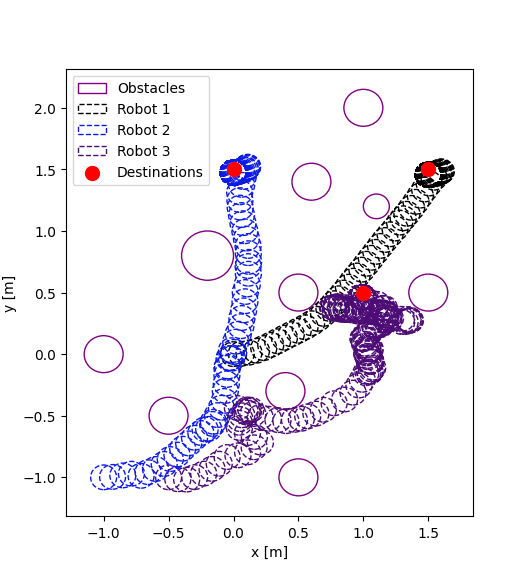}
\vspace{-12pt}
\caption{Qualitative results on target reaching in Map 1 for (left) nominal, (center) low-slip, and (right) high-slip settings. (Best viewed in color.)}\label{fig:tr_map1}
\vspace{-10pt}
\end{figure*}

We then validated the Koopman-based model learning when there was slip in the system. 
We set $\lambda_1 = 0$, $\lambda_2 = 0.1$, and $\lambda_3 = 0.2$ for the three robots to test how well the Koopman operator learned the dynamics with random slipping velocities. 
All other parameters were unchanged.  
The qualitative results are shown in Fig.~\ref{fig:model} (right). 
Although random slipping velocities exist, the Koopman operator can still learn smooth dynamic models from noisy data. 
While the patterns are generally followed, the numerical results show that the RMSE increases as the slip coefficient increases ($RMSE=[2.12\%,10.10\%,24.27\%]$ for robots $1,2,3$, respectively). 
We also assessed the Hausdorff distances between the predicted and true trajectories. 
Because the slip was randomly generated every time, this experiment was repeated 30 times, and the average Hausdorff distances were recorded at $[0.022m,0.068m,0.234m]$ for robots $1,2,3$, respectively.

\subsection{Koopman-NMPC for Target Reaching}\label{reaching}
Next, we evaluated our method for the target-reaching navigation problem using the models trained in~\ref{4.1}. 
Problem~\eqref{eq10} is posed symbolically via CasADi~\cite{andersson2019casadi}, discretized via multiple-shooting~\cite{leineweber2003efficient}, and solved via IPOPT~\cite{wachter2006implementation}. 
The control and state subspaces $\mathbb{U}$ and $\mathbb{X}$ in~\eqref{eq10} represent the constraints on the decision. 
Actuation limits are set at
$ -0.6\leq v_i  \leq 0.6$ and $-\pi/4  \le \omega_i  \leq  \pi/4$, for $i = 1,2,3$.
The robots are confined to operate within $(-10\leq (x_i,y_i) \leq 10)$.

The NMPC parameters are set as 
$Q=\begin{bmatrix}
Q_1 & 0 & 0 \\
0 & Q_2 & 0 \\
0 & 0 & Q_3\\
\end{bmatrix}$, 
with 
$Q_1=Q_2=Q_3=\begin{bmatrix}
4I_2 & 0\\
0 & 0.1I_2\\
\end{bmatrix}$, 
$R=\begin{bmatrix}
R_1 & 0 & 0 \\
0 & R_2 & 0 \\
0 & 0 & R_3\\
\end{bmatrix}$, with  
$R_1=R_2=R_3=\begin{bmatrix}
0.1 & 0\\
0 & 0.1\\
\end{bmatrix}$,
$T=20s$, and $t=100ms$.

Note that the NLP solver may be forced to violate the equality constraints. 
This, in turn, may produce robot velocities just over the actuation limits or lead to a collision with obstacles, even though the prediction model matches exactly how the system propagates. 
Adding a minimum distance constraint from obstacles can solve this problem; in our case, we set the offset at $0.015m$.

We test three cases: (1) the nominal setting (i.e. no slip at all), (2) a setting with a low amount of slip (slip in robots 2 and 3 given by $\lambda_2 = 0.05$ and $\lambda_3 = 0.1$), and (3) a setting with higher amount of slip ($\lambda_2 = 0.1$, $\lambda_3 = 0.2$). 
The first robot was assumed to be slip-free to test our method in a wider variety of settings simultaneously, given that we combined all robots into a larger optimization problem solved jointly. 
We deployed the system in maps (Map 1 and Map 2), each with a different distribution of obstacles and pairs of initial and target poses for each robot. 
In all cases, we performed ten trials. 

\begin{figure}[!t]
\vspace{-6pt}
\centering
\includegraphics[trim={0.5cm, 0.5cm, 1.5cm, 1.65cm},clip,width=0.49\textwidth]{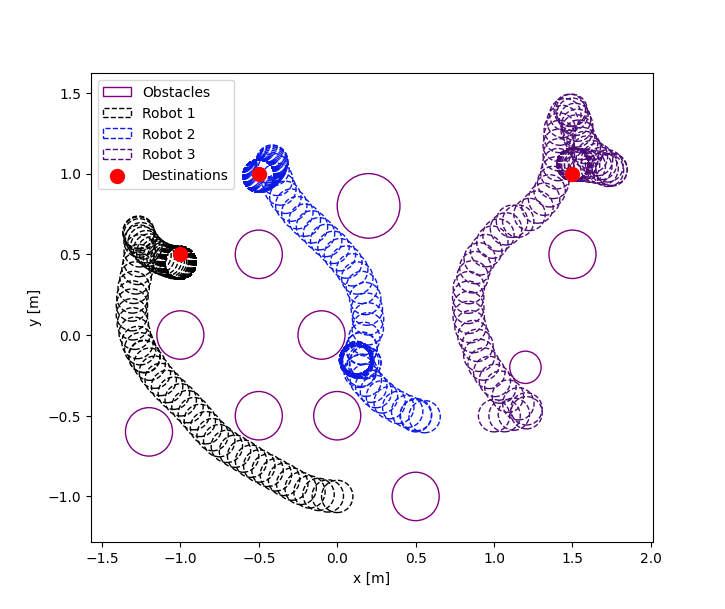}
\includegraphics[trim={0.5cm, 0.5cm, 1.5cm, 1.65cm},clip,width=0.49\textwidth]{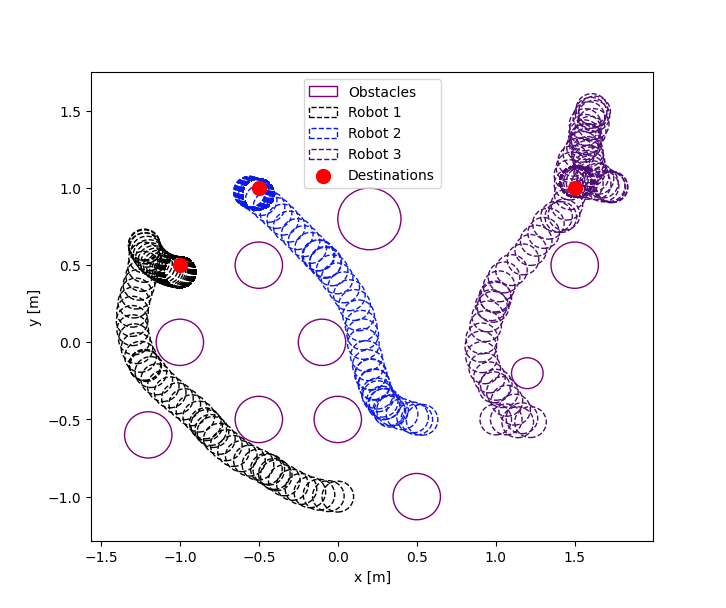}
\vspace{-15pt}
\caption{Sample target-reaching trajectories in Map 2 for (left) low-slip and (right) high-slip cases. The nominal case is not depicted for brevity but resembles the low-slip case, as in Map 1 (cf. Fig.~\ref{fig:tr_map1}). (Best viewed in color.)}\label{fig:tr_map2}
\vspace{-3pt}
\end{figure}

The qualitative results for Maps 1 and 2 are shown in Fig.~\ref{fig:tr_map1} and Fig.~\ref{fig:tr_map2}, respectively; Table~\ref{tab1} contains the quantitative results. 
The figures depict a sample trajectory for each case, while the table reports the success rates (SR) and final error in all trials for each case. 
The robots reached their desired targets in all cases under nominal settings, with no final position error exceeding $0.03m$. 
This shows that the Koopman model has learned the nominal system dynamics from simple trajectories (Section~\ref{sec3.1}) and the derived model can be paired well with our multi-robot NMPC framework.

\begin{table}[!t]
\centering
\caption{Koopman-NMPC Statistics for Target Reaching.}\label{tab1}
\vspace{-6pt}
\begin{tabular}{c c c c c c}
\toprule
\multirow{2}{*}{Map} & \multirow{2}{*}{Slip} & \multirow{2}{*}{SR} & \multicolumn{3}{c}{mean $\pm$ SD final distance [$\times10^{-2}m$]} \\
 &  &  & Robot 1 & Robot 2 & Robot 3\\
\midrule
\multirow{3}*{Map 1} & Nominal & 10/10 & $0.9$ & $1.9$ &$1.9$ \\
~ & Low & 9/10 & $3.1\pm0.7$ & $2.7\pm0.1$ & $6.6\pm4.1$ \\
~ & High & 8/10 & $1.9\pm1.5$ & $2.7\pm0.2$ & $6.6\pm1.9$ \\
\midrule
\multirow{3}*{Map 2} & Nominal & 10/10 & $2.8$ & $1.8$ & $2.3$ \\
~ & Low & 9/10 & $5.0\pm0.4$ & $2.4\pm0.4$ & $6.4\pm2.4$ \\
~ & High & 6/10 & $4.9\pm0.4$ & $2.6\pm0.7$ & $4.8\pm2.4$ \\
\bottomrule
\end{tabular}
\vspace{-20pt}
\end{table}

\begin{table}[!t]
\vspace{18pt}
\centering
\caption{Nominal NMPC Statistics for Target Reaching.}\label{tab2}
\vspace{-6pt}
\begin{tabular}{c c c c c c}
\toprule
\multirow{2}{*}{Map} & \multirow{2}{*}{Slip} & \multirow{2}{*}{SR} & \multicolumn{3}{c}{mean $\pm$ SD final distance [$\times10^{-2}m$]} \\
 &  &  & Robot 1 & Robot 2 & Robot 3\\
\midrule
\multirow{3}*{Map 1} & Nominal & 10/10 & $1.8$ & $1.5$ &$15.6$ \\
~ & Low & 8/10 & $1.3\pm0.4$ & $1.8\pm0.2$ & $3.5\pm2.7$ \\
~ & High & 3/10 & $1.1\pm0.4$ & $2.4\pm0.2$ & $4.0\pm2.5$ \\
\midrule
\multirow{3}*{Map 2} & Nominal & 0/10 & NaN & NaN & NaN \\
~ & Low & 1/10 & $2.4$ & $0.5$ & $3.4$ \\
~ & High & 0/10 & NaN & NaN & NaN \\
\bottomrule
\end{tabular}
\vspace{-5pt}
\end{table}

As uncertainty in the system increases, success rates drop (the lowest observed is $60\%$ and the average over all low- and high-slip settings and maps is $80\%$). 
The final position errors were more variable. 
Robot 3, which in all cases has the largest degree of slip, has a comparatively higher mean position error and larger dispersion (in terms of standard deviation SD), but overall, those numbers remain comparatively small (less than $10\%$) with respect to the overall trajectory length.
\looseness=-1\;\footnote{We observe that the final position of Robot 1 also varies slightly among trials. 
This is because the controller couples all robots together, which may create different controls for Robot 1 at different trials, depending on the current pose of the other robots in the team. Revising the optimizer can help address this, and is part of future work.}

To further demonstrate the effectiveness of the proposed method, we compare it with a nominal NMPC, where the prediction model is replaced by the nominal robot dynamics model~\eqref{eq16} without slip. 
All experiments were repeated under the same settings. 
Table~\ref{tab2} presents the statistics of the nominal NMPC. 
Results show that not only is the overall success rate lower than that of our proposed method, but also the overall final robot distances to their targets are much higher.

\begin{figure*}[!t]
\vspace{6pt}
\centering
\includegraphics[trim={0.5cm, 0.49cm, 1.75cm, 1.65cm},clip,width=0.327\textwidth]{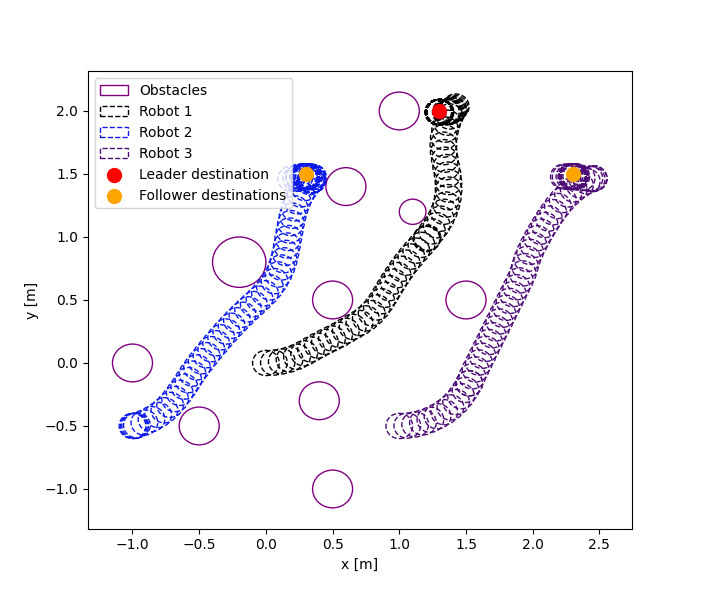}
\includegraphics[trim={0.05cm, 0.49cm, 1.75cm, 1.65cm},clip,width=0.327\textwidth]{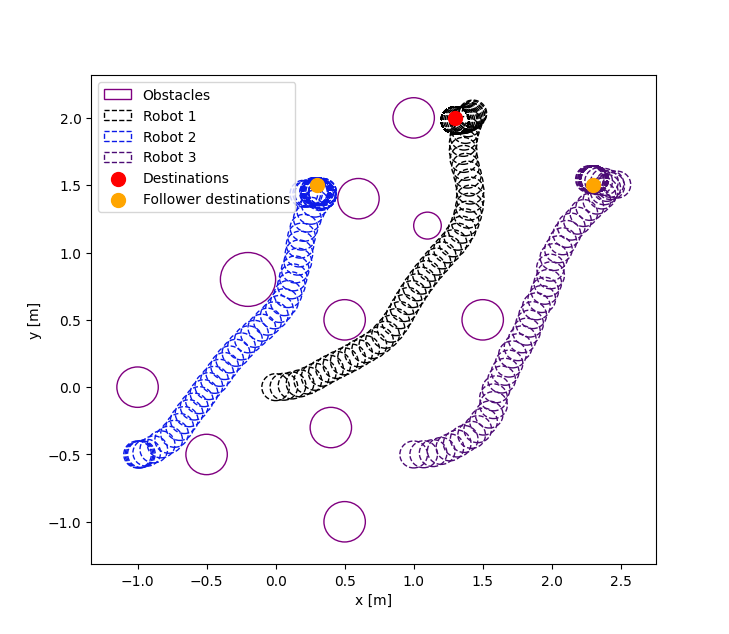}
\includegraphics[trim={0.05cm, 0.49cm, 1.75cm, 1.65cm},clip,width=0.327\textwidth]{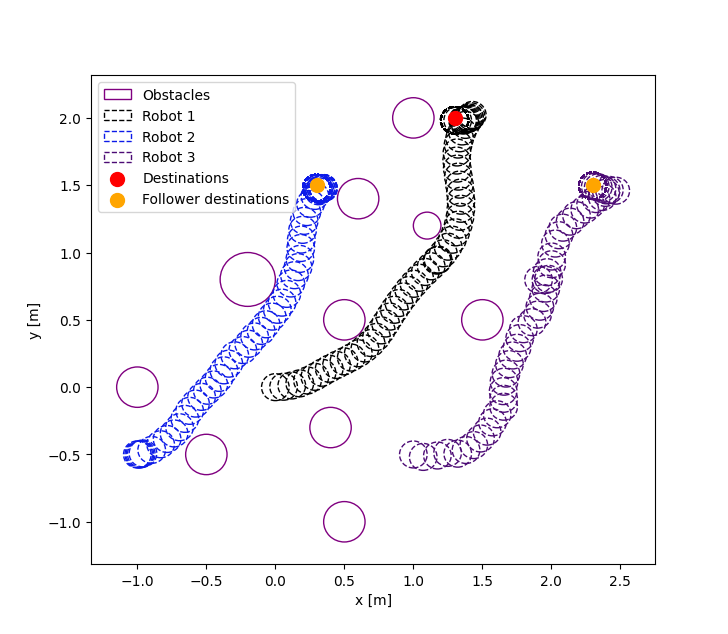}
\vspace{-25pt}
\caption{Qualitative results on formation control in Map 3 for (left) nominal, (center) low-slip, and (right) high-slip settings. (Best viewed in color.)}\label{fig:fc_map1}
\vspace{-12pt}
\end{figure*}

\subsection{Koopman-NMPC for Formation Control}\label{7}
To evaluate our method for formation control, we again used three robots and tested them on two maps. 
Map 3 has the same obstacle distribution as Map 1 before but with different initial robot poses and designated destination for the leader, while Map 4 is different. 
Robot 1 was selected as the leader. 
To capture its ``leading'' role in the optimization, $Q_2$ and $Q_3$ are revised to $\begin{bmatrix}
I_2 & 0 \\
0 & 0.1I_2 \\
\end{bmatrix}$. 
This reduces the impact of the follower robots on the objective function. 
All other parameters and settings remained the same as those in the target-reaching problem described previously.

\begin{table}[!t]
\vspace{0pt}
\centering
\caption{Koopman-NMPC Statistics for Formation Control.}\label{tab3}
\vspace{-0pt}
\begin{tabular}{c c c c c c}
\toprule
\multirow{2}{*}{Map} & \multirow{2}{*}{Slip} & \multirow{2}{*}{SR} & \multicolumn{3}{c}{mean $\pm$ SD final distance [$\times10^{-2}m$]} \\
 &  &  & Robot 1 & Robot 2 & Robot 3\\
\midrule
\multirow{3}*{Map 3} & Nominal & 10/10 & $0.9$ & $1.9$ &$1.9$ \\
~ & Low & 9/10 & $1.7\pm0.1$ & $4.5\pm0.1$ & $3.9\pm0.8$ \\
~ & High & 8/10 & $1.7\pm0.2$ & $2.4\pm1.1$ & $1.1\pm0.7$ \\
\midrule
\multirow{3}*{Map 4} & Nominal & 10/10 & $2.5$ & $7.0$ & $7.5$ \\
~ & Low & 8/10 & $2.6\pm0.0$ & $8.4\pm0.0$ & $9.6\pm4.0$ \\
~ & High & 8/10 & $3.0\pm0.0$ & $6.6\pm0.0$ & $13.5\pm0.0$ \\
\bottomrule
\end{tabular}
\vspace{-0pt}
\end{table}

\begin{figure}[!t]
\vspace{0pt}
\centering
\includegraphics[trim={0.35cm, 0.65cm, 1.5cm, 1.95cm},clip,width=0.45\textwidth]{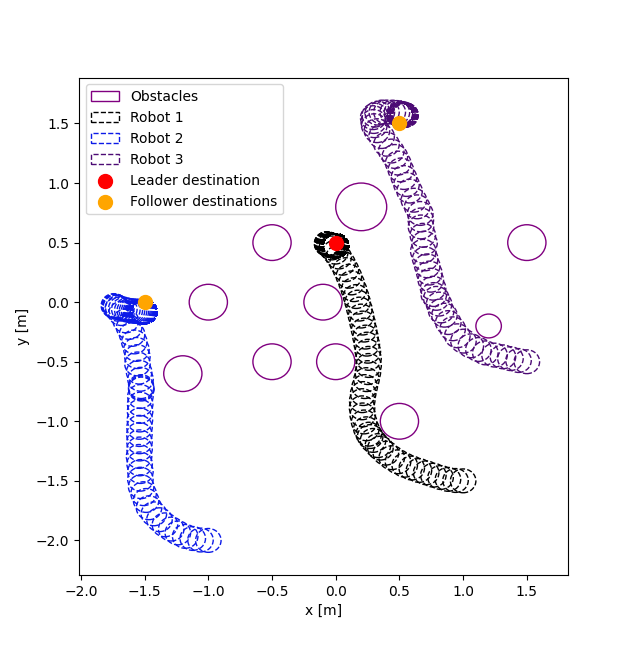}
\includegraphics[trim={0.35cm, 0.65cm, 1.5cm, 1.95cm},clip,width=0.45\textwidth]{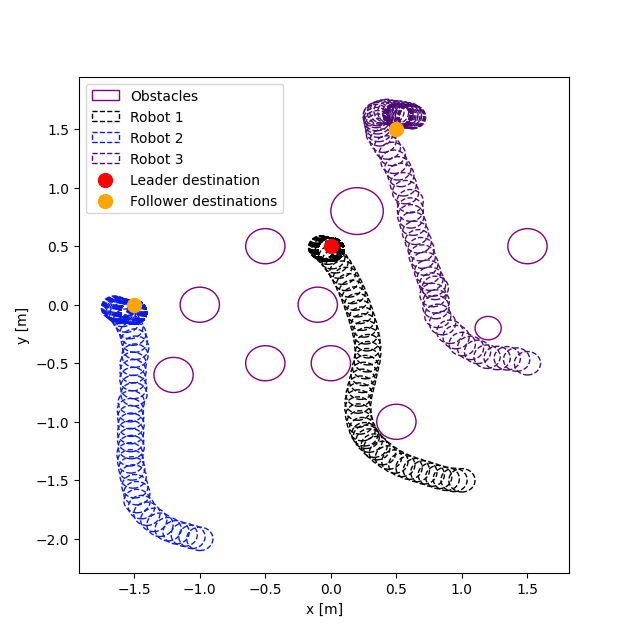}
\vspace{-10pt}
\caption{Sample formation control trajectories in Map 4 for (left) low-slip and (right) high-slip cases. The nominal case is not depicted for brevity but resembles the low-slip case, as in Map 3 (cf. Fig.~\ref{fig:fc_map1}). (Best viewed in color.)}\label{fig:fc_map2}
\vspace{-20pt}
\end{figure}

The qualitative results for Maps 3 and 4 are shown in Fig.~\ref{fig:fc_map1} and Fig.~\ref{fig:fc_map2}, respectively, while Table~\ref{tab3} contains quantitative results, similar to the target reaching problem presented previously. 
Figures depict a sample trajectory for each case while the tables report quantitative results regarding success rates and final error in all trials for each case. 
The robots successfully maintained their formation while the leader reached the designated destination in all cases under nominal settings. 
The final position errors for the follower robots were higher than those for the target-reaching cases. However, they do not exceed $0.08m$, which is less than $5\%$ of the total trajectory length (over $2m$).  
This finding further validates that the learned system dynamics can be applied to the more challenging case of formation control. 

As uncertainty in the system increases, the success rates decrease slightly but remain above $80\%$ in all cases. 
This suggests the efficacy of our Koopman-NMPC framework in solving the complex optimization problem of formation control. 
The mean position errors and their dispersion increased, similar to the target-reaching problem. 
One case (robot 3 in Map 3 and high-slip setting) appears to have a lower mean final error, but more trials would be needed to properly run statistical analysis and confirm that this is not due to chance. 
Nonetheless, the results were consistent and validated the efficacy of our method.
In contrast, the nominal NMPC fails in most cases (Table~\ref{tab4}). 

\begin{figure}[!t]
\vspace{0pt}
\centering
\includegraphics[trim={0.9cm, 1.7cm, 1.8cm, 3.0cm},clip,width=0.45\textwidth]{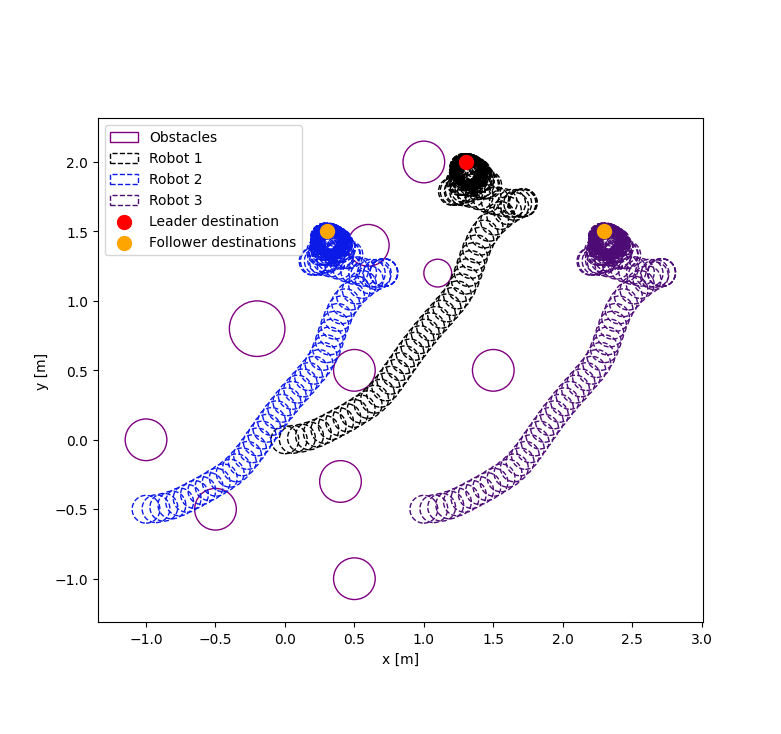}
\includegraphics[trim={1.5cm, 0.7cm, 2.6cm, 1.8cm},clip,width=0.43\textwidth]{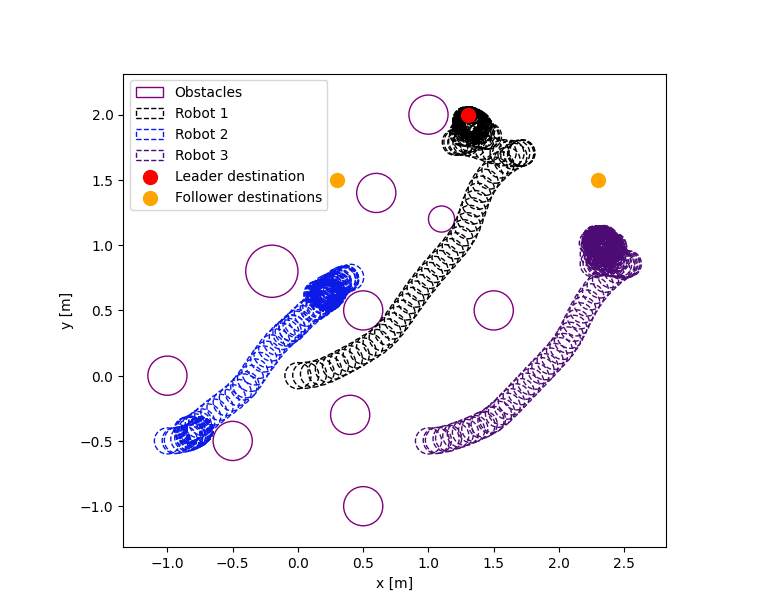}
\vspace{-10pt}
\caption{Sample formation control trajectories in Map 3 for the consensus-control method. (left) Ignoring collisions, consensus control provides well-matching trajectories. (right) When collision avoidance is
enforced, the followers fail to reach their targets within the allotted time, even in the nominal (no slip) case. 
(Best viewed in color.)}\label{fig:consensus}
\vspace{-12pt}
\end{figure}

Finally, we compared our method with the consensus-based multi-robot formation control~\cite{liu2020consensus}.
For fairness, we used the same maps and set the same maximum time constraints. 
First, we validated the method by testing it while ignoring collisions.
In this case, the consensus-based method generates appropriate trajectories, with the resulting trajectories of the followers matching those of the leader (an example shown in Fig.~\ref{fig:consensus}, left). 
However, when collision avoidance is considered, the consensus-based method cannot drive the follower robots to their target within the allotted time, even in the nominal case (one example shown in Fig.~\ref{fig:consensus}, right). 
Deviations along the trajectory to ensure collision avoidance make it difficult to maintain the desired relative distance between the robots. 
Table~\ref{tab5} summarizes the performance of the proposed consensus-based method. 
While the success rates appear to be similar between our Koopman-NMPC method and consensus-based formation control (Table~\ref{tab3}), there are large distances between the follower robots and their targets. 
These findings demonstrate that consensus-based control is applicable in open-environment navigation, but its feasibility in obstacle-cluttered environments is challenged.

\begin{table}[!t]
\vspace{0pt}
\centering
\caption{Nominal NMPC Statistics for Formation Control.}\label{tab4}
\vspace{-5pt}
\begin{tabular}{c c c c c c}
\toprule
\multirow{2}{*}{Map} & \multirow{2}{*}{Slip} & \multirow{2}{*}{SR} & \multicolumn{3}{c}{mean $\pm$ SD final distance [$\times10^{-2}m$]} \\
 &  &  & Robot 1 & Robot 2 & Robot 3\\
\midrule
\multirow{3}*{Map 3} & Nominal & 10/10 & $0.5$ & $1.0$ &$1.1$ \\
~ & Low & 0/10 & NaN & NaN & NaN \\
~ & High & 0/10 & NaN & NaN & NaN \\
\midrule
\multirow{3}*{Map 4} & Nominal & 10/10 & $1.7$ & $4.1$ & $3.8$ \\
~ & Low & 4/10 & $2.0\pm0.1$ & $3.7\pm0.5$ & $8.6\pm1.6$ \\
~ & High & 0/10 & NaN & NaN & NaN \\
\bottomrule
\end{tabular}
\vspace{-10pt}
\end{table}

\begin{table}[!t]
\vspace{0pt}
\centering
\caption{Consensus-based Statistics for Formation Control.}\label{tab5}
\vspace{-5pt}
\begin{tabular}{
>{\centering}p{1.0cm}  
>{\centering}p{1.50cm}
>{\centering}p{0.9cm}
>{\centering}p{1.3cm}
>{\centering}p{2.1cm}
>{\centering\arraybackslash}p{1.7cm}
}
\toprule
\multirow{2}{*}{~Map} & \multirow{2}{*}{Slip} & \multirow{2}{*}{SR} & \multicolumn{3}{c}{mean $\pm$ SD final distance [$\times10^{-2}m$]} \\
 &  &  & Robot 1 & Robot 2 & Robot 3\\
\midrule
\multirow{3}*{Map 3} & Nominal & 10/10 & $4.1$ & $76.3$ &$45.1$ \\
~ & Low & 9/10 & 4.1 & $148.3\pm 63.0$ & $50.3\pm4.5$ \\
~ & High & 9/10 & 4.1 & $158.3\pm54.9$ & $51.0\pm12.1$ \\
\midrule
\multirow{3}*{Map 4} & Nominal & 10/10 & $7.4$ & $71.4$ & $82.6$ \\
~ & Low & 10/10 & $7.4$ & $69.5\pm0.8$ & $113.2\pm66.2$ \\
~ & High & 7/10 & $7.4$ & $69.8\pm3.3$ & $90.9\pm73.7$ \\
\bottomrule
\end{tabular}
\vspace{-10pt}
\end{table}

\section{Conclusion}

In this paper, we developed and deployed a Koopman-based NMPC framework for the control of multi-robot systems. 
In contrast to previous related works that have considered learning only one follower robot in the leader-follower setup, our developed method learns well and predicts precisely the multi-robot system propagation under system uncertainty, which allows for runtime robot control at a high rate. 
Our method was tested without knowledge of the true dynamics of the system in both target reaching and formation control tasks for a 3-robot system in simulation with stochastic slip as the system uncertainty. 
The results demonstrated the feasibility of our method, and comparisons with other related methods revealed that our method can handle uncertainty well and achieve a high success rate in completing the given tasks. 
Notably, integrating a learned model via the Koopman operator theory can significantly increase the success rates of NMPC-based target reaching and formation control, while also leading to smaller endpoint distances compared with consensus-based control. 

This work lays the foundation for several future research directions. 
One aspect concerns the implementation and testing of physical hardware. 
Another direction for future research is the inclusion of more robots in the system and the formulation of other multi-robot navigation strategies.
Finally, the inclusion of control barrier functions could help enhance system safety, which is expected to become increasingly challenging as we move into physical hardware implementation with more robots in the team.

%
%
\bibliographystyle{plain}
\bibliography{ref}

\end{document}